\documentclass[journal]{IEEEtran}
\usepackage{ifpdf}
\usepackage{cite}
\usepackage{graphicx}
\usepackage[cmex10]{amsmath}
\usepackage{amsfonts}
\usepackage{algorithmic}
\usepackage{array}
\usepackage{fixltx2e}
\usepackage{stfloats}
\usepackage{url}
\usepackage{multirow}
\usepackage[backref]{hyperref}
\usepackage{subcaption}
\usepackage{color}
\usepackage{makecell}
\begin{document}
%
\title{Remote Sensing Object Counting with  Online Knowledge Learning}



\author{Shengqin Jiang, Yuan Gao, Bowen Li, Fengna Cheng, Renlong Hang,  Qingshan Liu,~\IEEEmembership{Senior Member, IEEE}


\thanks{Manuscript received *** **, 2024; revised *** **, 2024. This work is supported by the National Key Research and Development Program of China (No.2021ZD0112200), the Joint Funds of the National Natural Science Foundation of China (No.U21B2044), the National Natural Science Foundation of China (No.62001237), the Jiangsu Planned Projects for Postdoctoral Research Funds (No.2021K052A), the China Postdoctoral Science Foundation Funded Project (No. 2021M701756), the Startup Foundation for Introducing Talent of NUIST (No.2020r084). (\it{Shengqin Jiang, Yuan Gao and Bowen Li contributed equally to this work.}) (\it{Corresponding author: Qingshan Liu and Renlong Hang.})} 

\thanks{S. Jiang, Y. Gao, B. Li and R. Hang are with the School of Computer Science, Nanjing University of Information Science and Technology, Nanjing 210044, China, Ministry of Education Engineering Research Center of Digital Forensics, Nanjing University of Information Science and Technology, Nanjing  210044, China, and Jiangsu Collaborative Innovation Center of Atmospheric Environment and Equipment Technology (CICAEET), Nanjing University of Information Science and Technology, Nanjing 210044, China (e-mail: jiangshengmeng@126.com; gaoyuan\_mr@126.com; jslibowen@126.com; renlong\_hang@163.com)}
\thanks{F. Cheng is with the College of Mechanical and Electronic Engineering, Nanjing Forestry University, Nanjing, 210037, China (e-mail: cfn1218@163.com)} 
\thanks{Q. Liu is with the School of Computer Science, Nanjing University of Posts and Telecommunications, Nanjing, 210023, China (e-mail: qsliu@nuist.edu.cn).}
}

\markboth{IEEE Transactions on Geoscience and Remote Sensing}   
{Jiang \MakeLowercase{\textit{et al.}}:Remote Sensing Object Counting with  Online Knowledge Learning}

\maketitle

\begin{abstract}

Efficient models for remote sensing object counting are urgently required for applications in scenarios with limited computing resources, such as drones or embedded systems. A straightforward yet powerful technique to achieve this is knowledge distillation, which steers the learning of student networks by leveraging the experience of already-trained teacher networks. However, it faces a pair of challenges: Firstly, due to its two-stage training nature, a longer training period is essential, especially as the training samples increase. Secondly, despite the proficiency of teacher networks in transmitting assimilated knowledge, they tend to overlook the latent insights gained during their learning process. To address these challenges, we introduce an online distillation learning method for remote sensing object counting. It builds an end-to-end training framework that seamlessly integrates two distinct networks into a unified one. It comprises a shared shallow module, a teacher branch, and a student branch. The shared module serving as the foundation for both branches is dedicated to learning some primitive information. The teacher branch utilizes prior knowledge to reduce the difficulty of learning and guides the student branch in online learning. In parallel, the student branch achieves parameter reduction and rapid inference capabilities by means of channel reduction. This design empowers the student branch not only to receive privileged insights from the teacher branch but also to tap into the latent reservoir of knowledge held by the teacher branch during the learning process. Moreover, we propose a relation-in-relation distillation method that allows the student branch to effectively comprehend the evolution of the relationship of intra-layer teacher features among different inter-layer features. Extensive experiments on two challenging datasets demonstrate the effectiveness of our method, which achieves comparable performance to state-of-the-art methods despite using far fewer parameters.

\end{abstract}

\IEEEpeerreviewmaketitle




\section{Introduction}
\IEEEPARstart{R}{emote} sensing object counting involves  estimating the number of targets using images captured by satellites or drones. This technique is essential for applications such as resource management, urban planning, and environmental protection. Unlike general object counting, this task deals with large-scale images taken from high altitudes, where targets may be small, densely packed, and set against complex and variable backgrounds~\cite{gao2020counting, yi2023lightweight,xu2024rethinking}. These factors make robust object counting particularly challenging. Fortunately, recent rapid advancements in deep learning have significantly improved our ability to address these challenges.

\begin{figure}[!tb] 
	\centering
	\includegraphics[scale=0.55]{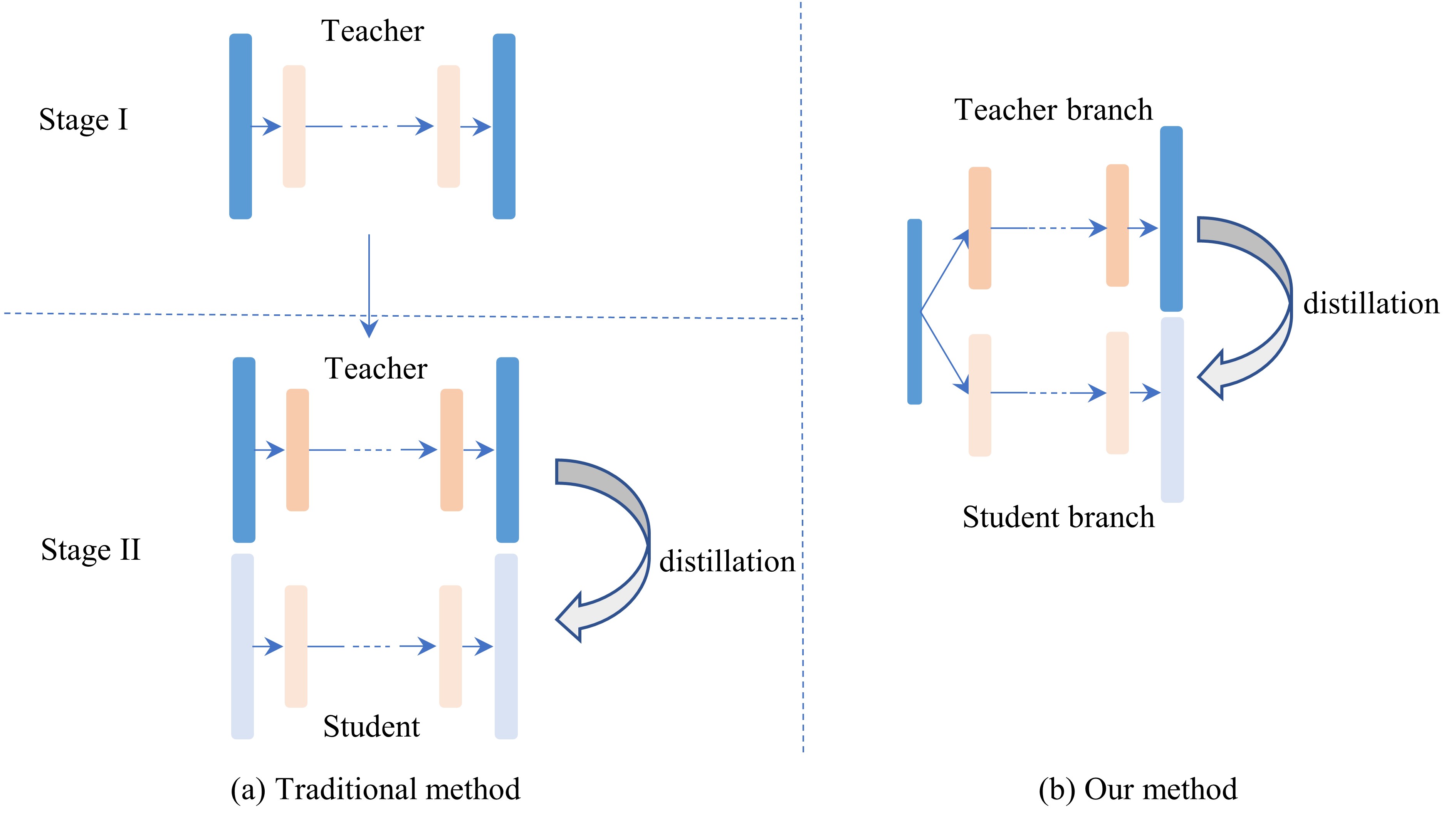}
	\caption{Comparison of distillation pipeline (traditional method vs. our method). (a) The traditional method involves training the teacher network first, followed by training the student network; (b) Our method employs an online distillation method to train the teacher and student branches jointly.}
	\label{fig:fig0}
\end{figure}

So far, considerable efforts have been devoted to the task of remote sensing object counting, yielding notable results. Most of these approaches achieve superior object counting performance through the careful design of specialized modules or loss functions. Gao et al.~\cite{gao2020counting} collected a large-scale dataset for remote sensing object counting and subsequently benchmarked it by designing a novel neural network based on the learning paradigm of MCNN~\cite{zhang2016single}. Duan et al.~\cite{duan2021distillation} fused context information from different receptive fields effectively and utilized feature maps from the deeper layer of the network to supervise feature maps from the earlier layer of the network. Yi et al.~\cite{yi2023lightweight} explored a multiscale feature fusion method to deal with challenges such as large-scale variation and complex background interference. Wang et al.~\cite{wang2024hierarchical} proposed several hierarchical kernel interaction modules to simultaneously preserve high-resolution features and extract deep-layer semantic information. These methods have established a crucial foundation for advancing remote sensing object counting. Upon further scrutiny, it becomes evident that current models predominantly adopt specific strategies built upon pre-trained networks. This strategy not only capitalizes on the powerful generalization capabilities of pre-trained models in downstream tasks but also allows for the customization of strategies tailored to the unique characteristics of remote sensing targets. While these models offer performance benefits, they typically exhibit a large number of parameters and high computational complexity. This makes them challenging to use in practical applications, such as drones or embedded systems, where processing resources are limited.



%
%
%

Numerous strategies have been proposed to develop small yet efficient models with comparable performance to larger models, such as pruning~\cite{cai2017deep, rao2018runtime} and quantization~\cite{park2018value}. Nonetheless, most of those methods require meticulous hyperparameter tuning or specialized hardware platforms~\cite{liu2020efficient,liu2020metadistiller}. An alternative approach, knowledge distillation, employs supervised information from a high-performing large model (referred to as the teacher) to train a smaller model (referred to as the student)~\cite{hinton2015distilling, zhang2018deep, zhang2019your, liu2020efficient}, as shown in Fig.~\ref{fig:fig0} (a). This method can substantially enhance the performance of the student and is both straightforward and practical to implement~\cite{dai2021general}. However, it has to face two challenges: 1) The method typically relies on two-stage training, which significantly increases the time required for training, particularly as the number of samples grows. 2) During the distillation process, although the student can learn rich semantic knowledge from the teacher network, it overlooks the insights gained by the teacher network during its own training.

To address the aforementioned challenges, we propose  an online knowledge learning method for remote sensing object counting. This framework, depicted in Fig.~\ref{fig:fig0} (b), is an end-to-end knowledge distillation approach, which avoids the need for two-stage training methods. It consists of a shared shallow module, a teacher branch, and a student branch. The student branch employs 1/4 of the channels of the teacher branch to reduce computational complexity. It is important to note that the pre-trained model initializes the shared shallow module and teacher branch to gain adequate domain-specific knowledge. Furthermore, we propose a relation-in-relation distillation strategy building upon feature distillation, which captures the evolution of intra-layer feature relationships across layers, thereby better capturing the intrinsic knowledge of the teacher branch. Finally, extensive experiments on two publicly available datasets validate the effectiveness of our method.

\begin{itemize}
\item To the best of our knowledge, we at the first time propose an end-to-end online knowledge distillation framework for remote sensing object counting. It utilizes a parallel dual-branch learning architecture to efficiently enhance the performance of the low-parameter student network while avoiding the time-consuming two-stage training process.
\item We propose a new relation-in-relation distillation method to assist the student branch in better understanding the evolution of intra-layer features by distilling the inter-layer relationship matrix from both branches.

\item Extensive experiments on two challenging datasets show the effectiveness of our framework. Moreover, our distilled model achieves comparable results to some state-of-the-art methods, despite having limited parameters.
\end{itemize}

\section{Related Works}

In this section, we will review some works related to our model. In what follows, the methods for remote sensing object counting will be introduced first. Then we will go through some recent efforts on knowledge distillation.

\subsection{Remote Sensing Object Counting}

In recent years, remote sensing object counting has garnered significant attention from both academia and industry~\cite{xia2018dota}. Early methods for object counting typically rely on detection-based approaches~\cite{wu2023yolo, mei2024scd}, which use object detection techniques to identify objects of interest in an image and predict bounding box information to determine the final count. Li et al.\cite{li2019simultaneously} introduced a unified framework that simultaneously detects and counts vehicles in drone images. Bayraktar et al.\cite{bayraktar2020low} employed YOLO V3 to detect filtered areas and locate target plants, subsequently estimating their number using geometrical relations and a predefined average plant size. Wu et al.~\cite{wu2023yolo} proposed a multiple-frequency feature fusion module and a bottleneck aggregation layer to enhance feature representations for counting smaller penguins. While detection-based approaches perform well in sparse scenarios, they may lead to significant inaccuracies when counting in highly dense or small-scale object settings.


A direct strategy involves designing a network to learn a mapping from input data to the number of people, which can be represented by a density map. This method simplifies the task of predicting bounding box coordinates. Gao et al.\cite{gao2020counting} developed an attention module, a scale pyramid module, and a deformable convolution module to address challenges in remote sensing object counting. They also collected a large-scale object counting dataset featuring four types of geographic objects. Ding et al.\cite{ding2022object} proposed an adaptive density map-assisted learning method to mitigate the loss of spatial features caused by groundtruth generated with fixed-size Gaussian kernels. To tackle issues such as scale variation and complex background interference, Yi et al.\cite{yi2023lightweight} explored a multiscale feature fusion method. More recently, Wang et al.\cite{wang2024hierarchical} addressed the loss of significant features in tiny-scale objects by sacrificing resolution to capture semantic information and designed a hierarchical kernel interaction network to resolve these issues. Guo et al.~\cite{guo2024balanced} introduced a balanced density regression network to reduce regression inaccuracies in Gaussian distributions caused by numerical variances.


The aforementioned methods typically build upon pre-trained models and then design specialized modules or strategies to address the challenges of the current task. As is well known, pre-trained models possess certain prior information, which allows them to quickly adapt to new scenarios. However, these models possess a substantial number of parameters, and incorporating additional modules can significantly elevate their computational resource requirements, thereby constraining their broader applicability. To address this, we explore a novel knowledge distillation framework that improves the training speed and enhances the transferability of privileged knowledge.

\subsection{Knowledge Distillation}

Knowledge distillation (KD) is a compression technique that transfers knowledge from a parameter-heavy pre-trained model to a more compact model. Hinton et al.\cite{hinton2015distilling} initially proposed using the output of a well-trained model as a supervisory signal to aid in the training of a student network. Romero et al.\cite{romero2014fitnets} utilized both the outputs and intermediate representations of the teacher model as cues to enhance the performance of the student model. Sau et al.\cite{sau2016deep} introduced a noise-based regularization method to strengthen the robustness of the student model. SKT\cite{liu2020efficient} leveraged the structured knowledge of a well-optimized teacher model to develop a lightweight student model. However, these strategies all require a time-consuming two-stage training process. Specifically, a high-capacity teacher model needs to be trained in advance before the student model can be effectively trained.


To sidestep this issue, online KD has been proposed to enable distillation within the network itself~\cite{yang2019snapshot, mirzadeh2020improved}. Yang et al.\cite{yang2019snapshot} utilized outputs from previous iterations as soft targets, but this approach risks introducing errors into the learning process and presents difficulties in selecting the appropriate iteration as a teacher. Zhang et al.~\cite{zhang2019your} introduced a self-distillation framework wherein knowledge from deeper layers assists in guiding the learning of shallower layers. Mirzadeh et al.~\cite{mirzadeh2020improved} proposed a multi-step KD method that employs a teacher assistant to bridge the gap between student and teacher networks. {Recently, Sun et al.~\cite{sun2024logit} studied the negative effects of the shared temperature between teacher and student during KD, and proposed setting the temperature as the weighted standard deviation of logit. FreeKD~\cite{zhang2024freekd} argued that consecutive downsamplings in the spatial domain of teacher models hinder the transfer of rich knowledge to student models. To address this, they introduced a frequency prompt to generate pixel-wise imitation signals and a channel-wise position-aware relational loss to improve the model's sensitivity to objects.} While these approaches reduce the training duration required for two-stage KD, they do not consider inter-feature information. Moreover, these studies primarily focus on relatively straightforward classification tasks, rather than addressing the more complex challenges associated with high-density prediction.




%

\begin{figure*}[!t]
	\centering
	\includegraphics[scale=0.6]{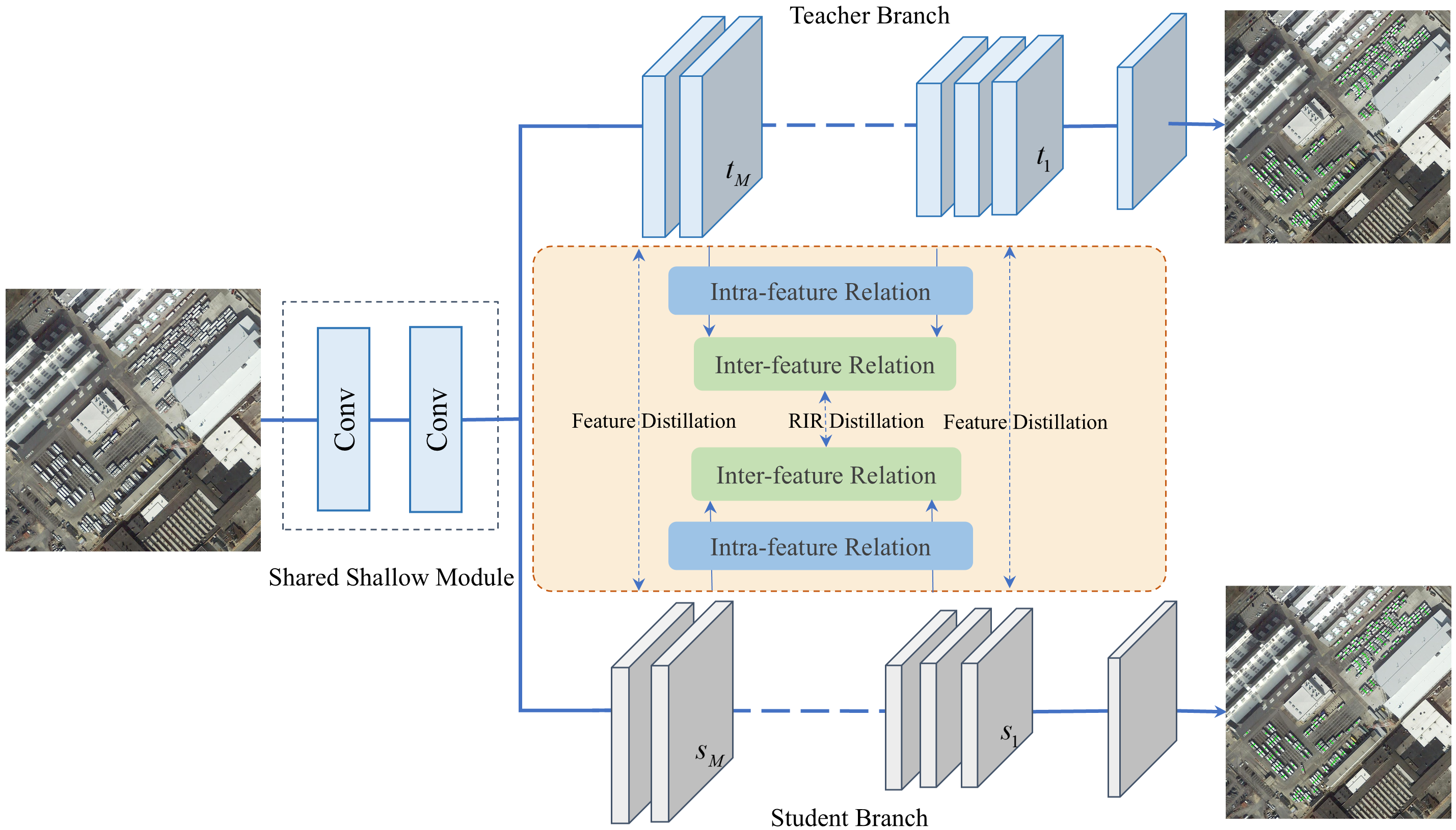}
	\caption{An overview of the proposed online distillation network. It consists of the shared shallow module, teacher branch and student branch. Meanwhile, feature distillation and relation-in-relation distillation are introduced to guide knowledge transfer from the teacher branch to the student branch.}
	\label{fig:fig1}
\end{figure*}

\section{Method}

{In this section, we present a novel Online Knowledge Learning method, termed OnKL Net, designed for remote sensing object counting.} Unlike the two-stage distillation training paradigm, our method is a one-stage training framework that can effectively transfer the inherent knowledge from the teacher branch, which holds privileged information, to the student branch. To accomplish this, we employ the point-based framework P2PNet~\cite{song2021rethinking} as our network foundation, which directly predicts individual locations as points. In what follows, we will begin by presenting our network architecture and then proceed to explain how we distill knowledge from the teacher branch to the student branch.


\subsection{Network Architecture}

Fig.~\ref{fig:fig1} illustrates the overall network structure, which is composed of a shared shallow module, a teacher branch, and a student branch. {As previously mentioned, we adopt a point-based counting framework that generates a set of potential point prompts through a regression head~\cite{song2021rethinking}. The optimal target positions are then refined using a classification head, which estimates the probabilities of these prompts. In particular, the input image is initially processed by the shared shallow module to extract base features before being forwarded to the two branches. Both the teacher branch and the student branch predict a set of point proposals and their corresponding confidence scores,  with the former serving as a mentor to the latter during training. To ensure the learning objectives of these proposals, we use a one-to-one matching between point proposals and ground truth points.} 


\subsubsection{Shared Shallow Module}

The early layers of 2D CNNs are commonly used to extract low-level features such as edges and corners. Here, we utilize a shared two-layer convolutional module for extracting spatial patterns. Specifically, we adopt the first two layers of VGG-16~\cite{simonyan2014very} for this purpose, as they are effective and widely used in various tasks. This module consists of two $3\times 3$ convolutions followed by max-pooling with a stride of (2, 2).

Since the teacher branch requires prior knowledge (please refer to the subsection~\ref{teabranch} for details), the starting point of the module is to keep the network structure consistent with the shallow layer of the pre-trained model. In light of this, we initialize this module using the pre-trained parameters. It is important to note that the shared module also serves as the initial layers of the student branch, so it should not have too many parameters. In other words, we should avoid introducing too many shallow layers in the teacher branch when designing the shared module.



\subsubsection{Teacher Branch} \label{teabranch}

In addition to using the first two layers of VGG-16 in the shared shallow module, we employ the remaining layers of the first 10 layers of VGG-16 as the front-end of the teacher branch. This allows us to utilize the pre-trained model, which has strong transfer learning capabilities and a flexible design. The teacher branch is intended to act as an experienced mentor who can quickly learn new information and guide the student branch.

\subsubsection{Student Branch}


The channel capacity of a network is a critical factor that affects both inference speed and memory consumption. {In our framework, we restrict the number of channels in the student branch to one-fourth of those in the teacher branch to reduce the computational burden of the student branch.} However, following the design principles of the VGG-16 network, directly reducing the number of channels may limit the network's feature representation capability.

To address this issue, we adopt a compromise method in which we keep the number of channels in the shared shallow module constant until the first pooling layer. After that, we reduce the number of channels in the remaining layers by one-fourth. Additionally, we can further reduce the number of channels by using powers of one-half. This approach allows us to strike a balance between channel capacity and computational efficiency while maintaining the expressive power of the network.

\subsection{Relation-in-Relation Distillation}


The limited number of channels in the student branch can hinder its ability to extract features effectively. Fortunately, feature distillation offers an effective solution to this issue by providing additional supervisory signals for knowledge transfer~\cite{li2020local, gou2021knowledge, heo2019knowledge}. This method can allow the student network to capture the representational details of the teacher network at different levels but overlooks how the teacher model combines different features to form higher-level abstractions. To this end, we propose a relation-in-relation distillation strategy that enables the student network to effectively learn the associative information embedded between features in the teacher network. Before delving into the details of these techniques, we introduce a feature grouping approach to better understand our proposed knowledge distillation strategy.

\subsubsection{Feature Grouping}

In Fig.~\ref{fig:fig1}, an input image is first processed by the shared shallow module, which outputs a set of low-level features. These features are then fed into two separate branches with different channel capacities. Let us denote these features as $t_{base} (s_{base})$ to represent the outputs of the shared shallow module. To facilitate clarity and simplify analysis, we sequentially number the feature outputs from the deeper layers to the shallower layers in the network. The group partition begins with the convolution following shallow pooling operations and ends just before deep pooling operations. The student branch follows the same convention. We define the feature output of the deepest group in the teacher (student) branch as $t_1$ ($s_1 $). Formally, the features are grouped as follows:



\begin{equation}
T{\text{ = }}\{ {t_1},\;{t_2},\;...,{t_M}\} ,\;S{\text{ = }}\{ {s_1},\;{s_2},\;...,\;{s_M}\} ,
\end{equation}\\
where $T$ and $S$ denote the feature sets of the teacher branch and the student branch, respectively. $M$ denotes the number of feature groups.

The features cannot be distilled directly since the number of channels in the two branches is not aligned. Therefore, to maintain consistency in the feature dimensions, we apply a linear transformation function $f_{ad}(\cdot)$ to the output features of various groups of the student network. This transformation function adapts $s_i$ to $s_i^\prime$, which has the same dimension as $t_i$. The transformation is defined as follows:


\begin{equation}
{S^\prime } = {f_{ad}}(S)  = \{ s_1^\prime ,\;s_2^\prime ,...,s_M^\prime \}, 
\end{equation}\\
where $S^\prime$ means the feature set obtained by transforming each element of $S$. It should be noted that this transformation is only employed during the training stage.

\subsubsection{Relation-in-Relation Distillation}

\begin{figure}[!t]
	\centering
	\includegraphics[scale=0.6]{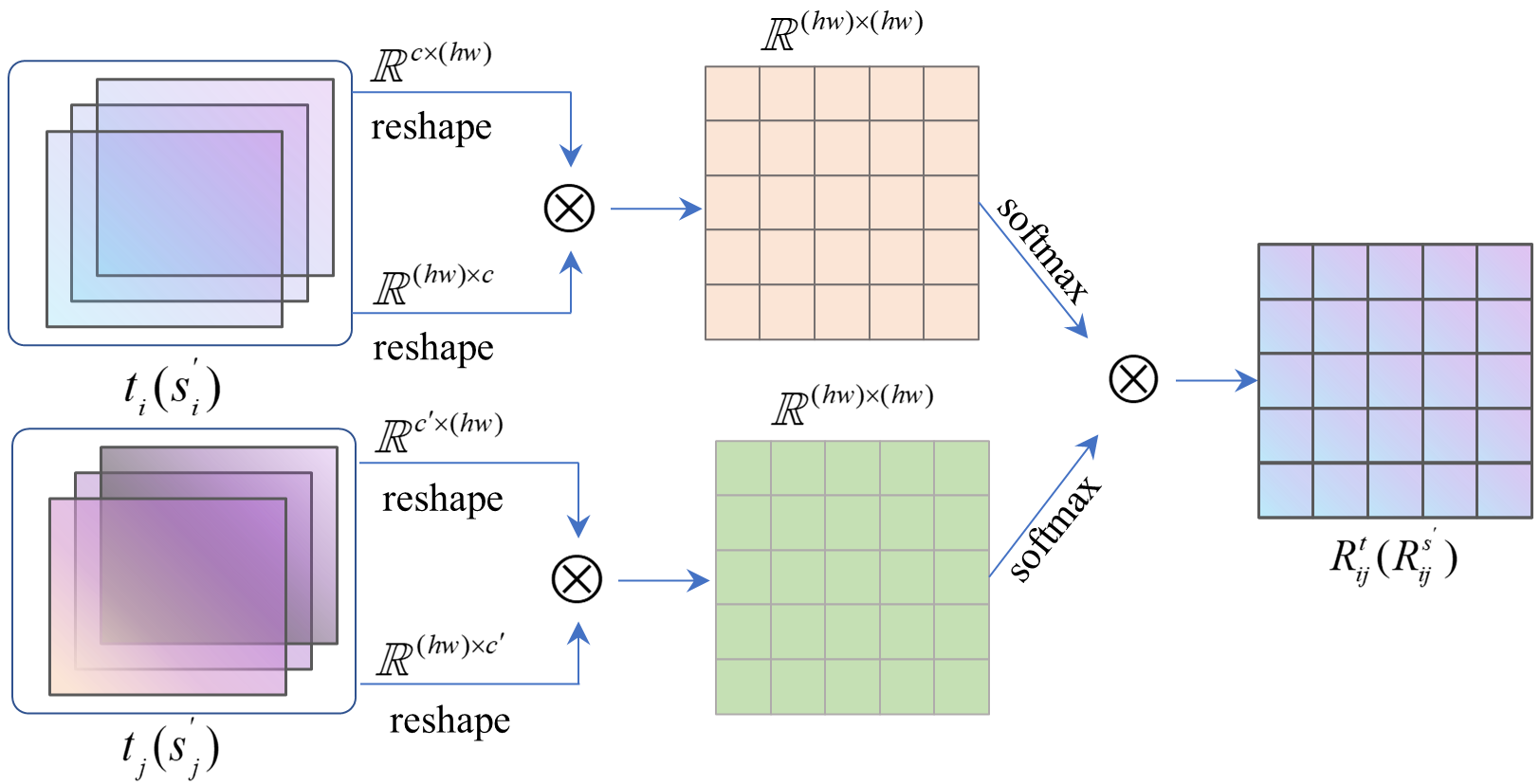}
	\caption{Overview of relation-in-relation feature transformation. {The inputs come from two different feature maps, $t_i (s_i^\prime)$ and $t_j (s_j^\prime)$, from different groups in the teacher (or student) branch, where $1 \le i < j \le M$.}}
	\label{fig:fig2}
\end{figure} 




{In deep networks, the ability to abstract intrinsic information in images increases layer by layer, reflecting the evolution of features within the network, which is referred to as the inter-layer "flow" in~\cite{yim2017gift}.} Therefore, exploring the relationships between layers is crucial for helping student networks capture robust features. For example, SKT~\cite{liu2020efficient} densely computed pairwise feature relationships using down-sampled block features.  Existing methods mainly focus on the relationships between individual feature pixels across different groups, which does not effectively capture the relationships among pixels within features of the same group. This makes it difficult to identify regions of interest with highly repetitive patterns in remote sensing scenes, particularly when the targets are small, have irregular shapes, or are similar to other objects. To address this issue, we propose a relation-in-relation distillation method which consists of a relation-in-relation feature transformation for features from different groups. The transformation models the relationships between different pixels (named intra-feature relation) and the corresponding similarity between layers (named inter-feature relation),  after which a distillation loss is applied to help the student branch  to learn the evolution of intra-feature relations between layers.


The overview of the relation-in-relation feature transformation is depicted in Fig.~\ref{fig:fig2}. To better elucidate the transformation, we  will utilize the teacher branch as an example. Firstly, we can obtain two different feature maps $t_i\in \mathbb{R}^{B\times C_1\times H\times W}$ and $t_j\in \mathbb{R}^{B\times C_2\times H\times W} (1\le i<j\le M)$ from different groups. Next, we apply a reshape operation to transform the shape of the two feature maps, for example, for $t_i$, we obtain two new feature maps: $t_i^1\in \mathbb{R}^{B\times (HW) \times C_1}$ and $t_i^2\in \mathbb{R}^{B\times C_1\times (HW)}$.  Multiplying $t_i^1$ by $t_i^2$, we obtain the correction matrix $K_i \in \mathbb{R}^{B\times HW\times HW}$ after applying the softmax operation $g( \cdot )$. Following the same operation, we derive the correction matrix $K_j$ for the feature map $t_j$. {To model the inter-layer relationships, we apply element-wise multiplication, which maintains the spatial structure of the intra-layer correlations while processing each position on the feature map independently. This allows the model to selectively enhance or suppress feature responses in specific regions.} Finally, the relation matrix between the $i$-th and $j$-th group in the teacher branch is obtained, denoted as $R_{ij}^t = K_i \otimes K_j$.  Formally, the inter-layer relationship matrix can be expressed as follows:
\begin{equation}\label{eq3}
R_{ij}^t = \frac{{g(t_i^1  t_i^2) \otimes g(t_j^1 t_j^2)}}{{HW}}
\end{equation} \\
where $\otimes$ denotes the element-wise product. After performing the same operation, we get the same relationship matrix of the student branch $R_{ij}^{s^\prime}$ for the $i$-th and $j$-th group.

Here we simply use the L1 distance to measure the gap between the corresponding relationship matrices. The relation-in-relation distillation loss is defined as follows:
\begin{equation}
	{L_r} = \sum\limits_{1 \le i < j \le M} {{{\left\| {R_{ij}^{{s^\prime }} - R_{ij}^t} \right\|}_1}}.
\end{equation}

\subsection{Total Loss}

To optimize the network parameters, the overall optimization loss is given as follows:

\begin{equation} \label{loss}
	{L_{total}} =  {\alpha _1}{L_f} + {\alpha _2}{L_r} + {L^s }+ {L^t },
\end{equation} \\
where $L_f$ denotes the feature distillation loss, defined as follows:
\begin{equation}
L_f=\sum_{i=1}^M{\left\| t_i-s_{i}^{\prime} \right\| _{2}^{2}.}
\end{equation}
where ${\left\|  \cdot  \right\|_2}$ denotes the 2-norm. Here, we use  the mean squared error loss (MSE loss) to align features from the same group of teacher and student networks, where we apply $1\times 1$ convolutions to the features of the student branch to match the number of channels with that of the teacher branch.  $\alpha_1$ and $\alpha_2$ are tunable hyper-parameters to balance the loss terms. ${L^s }$ and ${L^t }$ denote the supervision signals of two branches, respectively. {For each branch, following the optimization loss in P2PNet~\cite{song2021rethinking}, we employ the cross-entropy loss $L_{cls}^\ell$ and the MSE loss $L_{loc}^\ell $ to supervise the classification head and regression head, respectively, where $\ell = s, t$. Formally, the final loss ${L^\ell }$  is defined as follows:}
\begin{equation}
\begin{array}{l}
	{L^\ell } = L_{cls}^\ell  + \vec \lambda L_{loc}^\ell ,\\
	L_{cls}^{\ell}=-\frac{1}{V}\left( \sum_{i=1}^U{\log \widehat{c}_{\xi (i)}}+\hat{\lambda}\sum_{i=U+1}^V{\log\mathrm{(}1-\widehat{c}_{\xi (i)})} \right), \\
	L_{loc}^{\ell}=\frac{1}{U}\sum_{i=1}^U{\left\| p_i-\widehat{p}_{\xi (i)} \right\| _{2}^{2},}
\end{array}
\end{equation}
{where $U$ and $V$ denote the number of positive and total proposals, respectively,  ${p_i}$ represents the head’s center point of the individual, ${\widehat c_{\xi (i)}}$ is the confidence score of the predicted point ${\widehat p_{\xi (i)}}$, $\hat \lambda $  is a weight factor for negative proposals, and $\vec \lambda$  is a weight term to balance the effect of the regression loss.}

\section{Experiment}

In this section, we conduct extensive experiments on two challenging datasets to illustrate the effectiveness of our proposed strategy.

\subsection{Experiment Settings}


All experiments in this study are implemented using PyTorch. Data augmentation is performed as follows: The image is randomly scaled to 0.7 to 1.3 times its original size. The scaled image is then randomly cropped into four 128 $\times$ 128 patches. Next, random flipping is applied. VGG-16\_bn pretrained on ImageNet is used to initialize the first ten layers of the shared shallow module and the teacher branch. The remaining layers are initialized by a random Gaussian distribution with a standard deviation of 0.01. For training, the initial learning rate of the student branch and teacher branch is set to $1e^{-4}$. The same hyper-parameter settings $( \alpha_1=0.5, \alpha_2=5e^{-3} )$ are adopted for all experiments. The network is trained for 1500 epochs on the building dataset and 3000 epochs on the remaining datasets. The Adam optimizer is utilized to optimize the network parameters. 

{To ensure a fair evaluation of  counting methods, we use the Mean Absolute Error (MAE) and Root Mean Squared Error (RMSE) as evaluation metrics. Additionally, we characterize the model's computational complexity using training time (Time), the number of parameters (Param), and Floating Point Operations Per second (FLOPs), with the input resolution set to $256 \times 256$. Note that Time denotes the training time of models on Large-vehicle.}

\subsection{Datasets}

 RSOC is a publicly available dataset~\cite{gao2020counting} used to evaluate the counting performance of various classes of remote sensing objects. It includes 280 images of small vehicles, 172 of large vehicles, 137 of ships, and 2,468 images of buildings, totaling 3,057 images.  The dataset excludes easy cases with disperse objects, focusing on challenging scenarios like crowded ships near shores and densely packed vehicles in parking lots. For a fair evaluation of our method, we partition the dataset as follows for each category:  Small-vehicle includes 200 images for training, 22 for validation, and 58 for testing;  Large-vehicle has 98 training images, 10 validation images, and 64 testing images;  Ship consists of 88 training images, 9 validation images, and 40 testing images; and Building, which has the highest count, features 964 training images, 241 validation images, and 1,263 testing images.

{STAR~\cite{li2024star} is introduced for scene graph generation in large-scale high-resolution satellite imagery. The dataset includes over 210,000 annotated objects, with an emphasis on diverse scenarios. We select five typical object categories from this dataset as analysis objects: airplanes, windmills, lattice towers, bridges, and runways.  Specifically, for the airplane category, 233 images are selected for training and 106 for validation. The windmill category includes 29 images for training and 11 for validation. For lattice towers, there are 132 images for training and 39 for validation. In the case of large-scale objects, the bridge category contains 198 images for training and 43 for validation, while the runway category includes 239 images for training and 106 for validation.}

\subsection{Ablation Study}
\begin{table*}
	\caption{Comparison of two-stage distillation and online distillation. {The parameters provided here refer exclusively to those used during the inference phase of the model. }}
	\label{tab:two phase}
	\centering
	\begin{tabular}{ccccc}
		\hline
		Method                                             & Param(M)     & Time (min)     & MAE        &RMSE     \\
		\hline    
		SKT~\cite{liu2020efficient}                        &  1.02         &   459     &42.43      &   68.42                \\
		\hline
		Two-stage Distillation (Ours)                     & 0.76       &   141       &21.97	& 50.57                    \\
		OnKL Net  w/o pre-trained VGG-16            &0.76     & 85   &  14.58		&24.09                       \\
		OnKL Net                                       &  0.76         &  85         &\textbf{12.27}  &\textbf{16.70}             \\
		\hline
	\end{tabular}
\end{table*}

In this part, we conduct  ablation studies on Large-vehicle to assess the effectiveness of our proposed method. As mentioned in~\cite{liu2020efficient}, the number of channels is an important factor in balancing efficiency and accuracy in knowledge distillation. Therefore, we follow its approach and set the channel number of the student branch to be one-fourth of the teacher branch. We provide a detailed analysis of our ablation study in the following subsections.

\paragraph{Exploration on Online Knowledge Learning}\label{online}
We present the experimental results to evaluate the effect of online knowledge learning. Table~\ref{tab:two phase} shows the comparison between two-stage distillation and online distillation. Our proposed online distillation method outperforms the two-stage distillation method by a large margin, demonstrating the benefits of joint training. Specifically, our online distillation achieves a performance gain of more than 44\% over the two-stage distillation (Ours) in terms of MAE and RMSE. Moreover, compared to the two-stage version, the online version only requires about 60\% of its training time, even with only 98 training samples. This means that the time discrepancy between the two methods can become increasingly apparent as the sample size grows. To further validate the superiority of our method, we compare our network with different baselines in terms of inference speed. Here, all experiments are tested on an NVIDIA GeForce RTX 3090, with the test images having a resolution of $512 \times 512$. Our inference speed reaches 319.9 frames per second (FPS), which is superior to P2PNet$^{*}$'s 111.3 FPS and P2PNet's 88.6 FPS. Thereinto,  P2PNet$^{*}$ is a simplified version of P2PNet, with reduced channel numbers in both its regression and classification branches.

Our proposed solution of online distillation, with only 0.75 times the parameters of the student network in SKT~\cite{liu2020efficient}, is greatly better than this two-stage method (12.27 (16.70) vs. 42.43 (68.42)). These results demonstrate the efficacy of our distillation method as well as its high potential for knowledge transfer. Furthermore, our final solution initializes with the parameters of a pre-trained VGG-16 model, resulting in a slight performance improvement compared to the model without it (12.27(16.70) vs. 14.58(24.09)). This demonstrates that the teacher branch with prior knowledge can better support the student branch to capture effective knowledge. {We observe that the version of OnKL Net without a pre-trained model outperforms the two-stage distillation version. We speculate that this may be due to the current distillation strategies, which fail to effectively transfer knowledge from the teacher model, hindering the network's ability to achieve the desired results. In comparison, OnKL Net, which is also initialized by a pre-trained model, achieves even more superior performance. This further demonstrates the effectiveness of the  distillation strategies within our distillation framework.}


\paragraph{Exploration on Knowledge Distillation Configurations}

To achieve efficient knowledge transfer, we employ two knowledge distillation strategies: feature distillation (FD) and relation-in-relation distillation (RiRD). We establish a baseline without any distillation strategies to evaluate the effectiveness of different losses. Table~\ref{tab:tab1} presents the ablation results for the two distillation losses. As shown in the table, both loss functions significantly enhance the performance of the student network, demonstrating the effectiveness of these methods. The combination of these two losses further improves performance, yielding an MAE of 12.27 and an RMSE of 16.70, which suggests that these losses are complementary.

\begin{table}
	\caption{Ablation study of knowledge distillation strategies.}
	\label{tab:tab1}
	\centering
	\begin{tabular}{ccccc}
		\hline
		Module         & FD loss     & RiRD loss       & MAE           & RMSE \\
		\hline
		Baseline       &             &                & 19.64         &37.78  \\
		\hline        
		               & \checkmark  &                &   13.59       &18.73  \\
		               &             &  \checkmark    &15.42          &24.29  \\
		               & \checkmark  &  \checkmark    &  \textbf{12.27}     &\textbf{16.70}\\
		\hline
	\end{tabular}
\end{table}

\begin{table}
	\caption{Comparison of different feature relation distillation losses. {RiRD loss denotes the relation-in-relation distillation loss.}}
	\label{tab:tab3}
	\centering
	\begin{tabular}{ccc}
		\hline
		Transfer Configuration       & MAE  &RMSE\\
		\hline
		W/O RiRD loss             &13.59  &18.73\\
		\hline
		+ FSP loss~\cite{liu2020efficient}     &14.375  &20.92\\
		
		+ Cos loss                             &14.73 &19.65\\
		
		
		+ RiRD loss                              &\textbf{12.27}  &\textbf{16.70}\\
		\hline
	\end{tabular}
\end{table}

\paragraph{Effects of Relation-in-Relation Distillation} 

Table~\ref{tab:tab3} compares the performance of our proposed relation-in-relation distillation with other strategies. Our results show superior performance. It is notable that incorporating both FSP loss~\cite{liu2020efficient} and Cos loss to measure relationships between features in different groups results in lower performance compared to using only feature distillation. We speculate that this may be due to the highly similar patterns in remote sensing images, which complicate the task of mining feature relationships and may even degrade performance. Additionally, FSP is designed for distillation from a well-trained teacher network, whereas the features of the teacher branch in our network are dynamically changing, leading to weaker pixel-level feature correlations. This suggests that our method is effective in handling such complex scenarios.


\begin{figure}[htbp]
	\centering
	\includegraphics[scale=0.3]{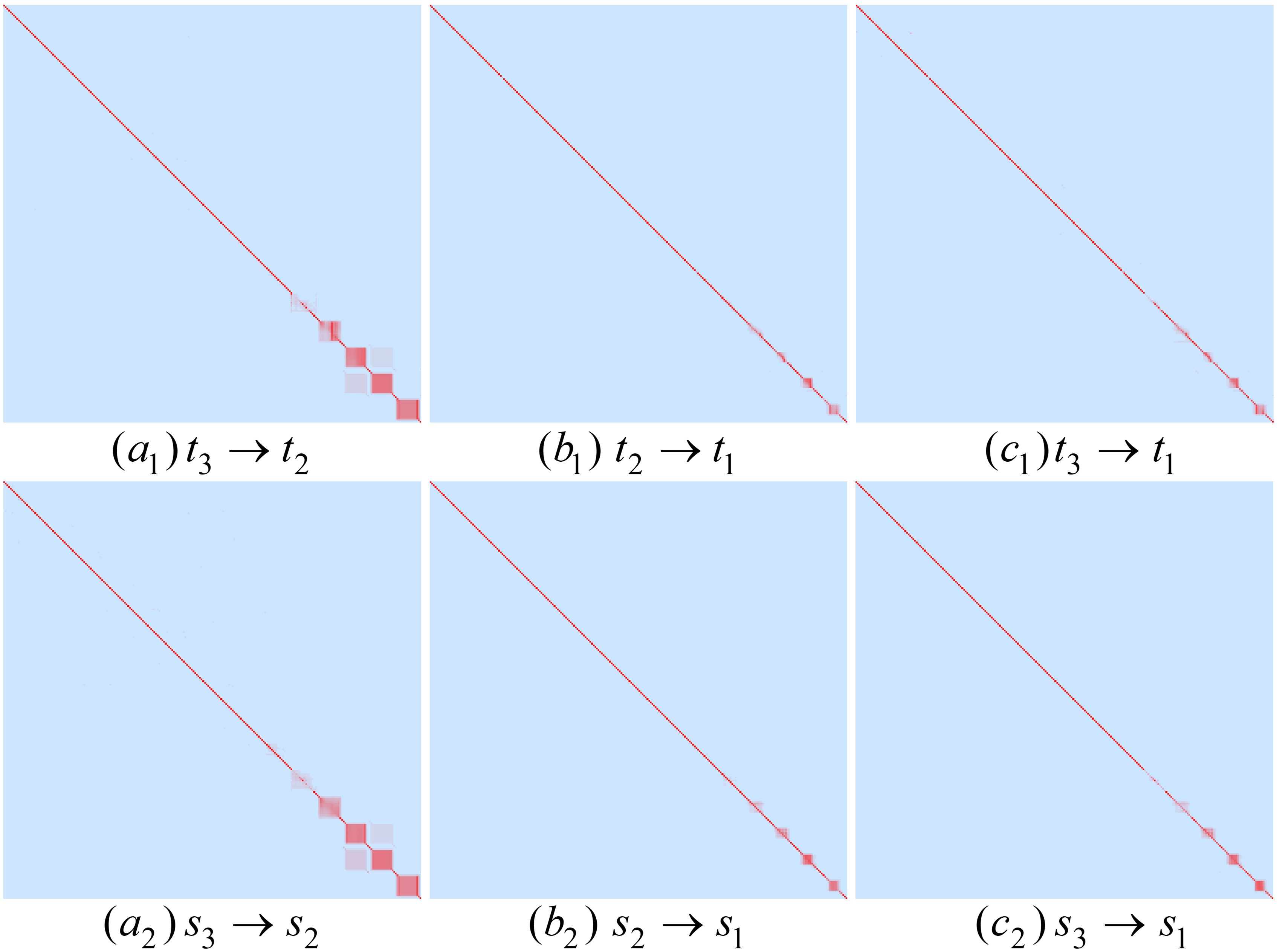}
	\caption{Visualization of  the inter-layer relationship matrix. The relationship matrices of the teacher and student branches are represented in the first and second rows, respectively. ${t_i} \rightarrow {t_j} ({s_i} \rightarrow {s_j})$ denotes the relation matrix generated by ${t_i} ({s_i})$ and ${t_j}  ({s_j})$. To clearly highlight between strong and weak relevant relationships, we apply the transformation $-1/log(x)$ to the non-zero values of the relationship matrix.}
	\label{fig:fig4}
\end{figure}

\begin{figure*}[ht] 
	\centering
	\includegraphics[scale = 0.55]{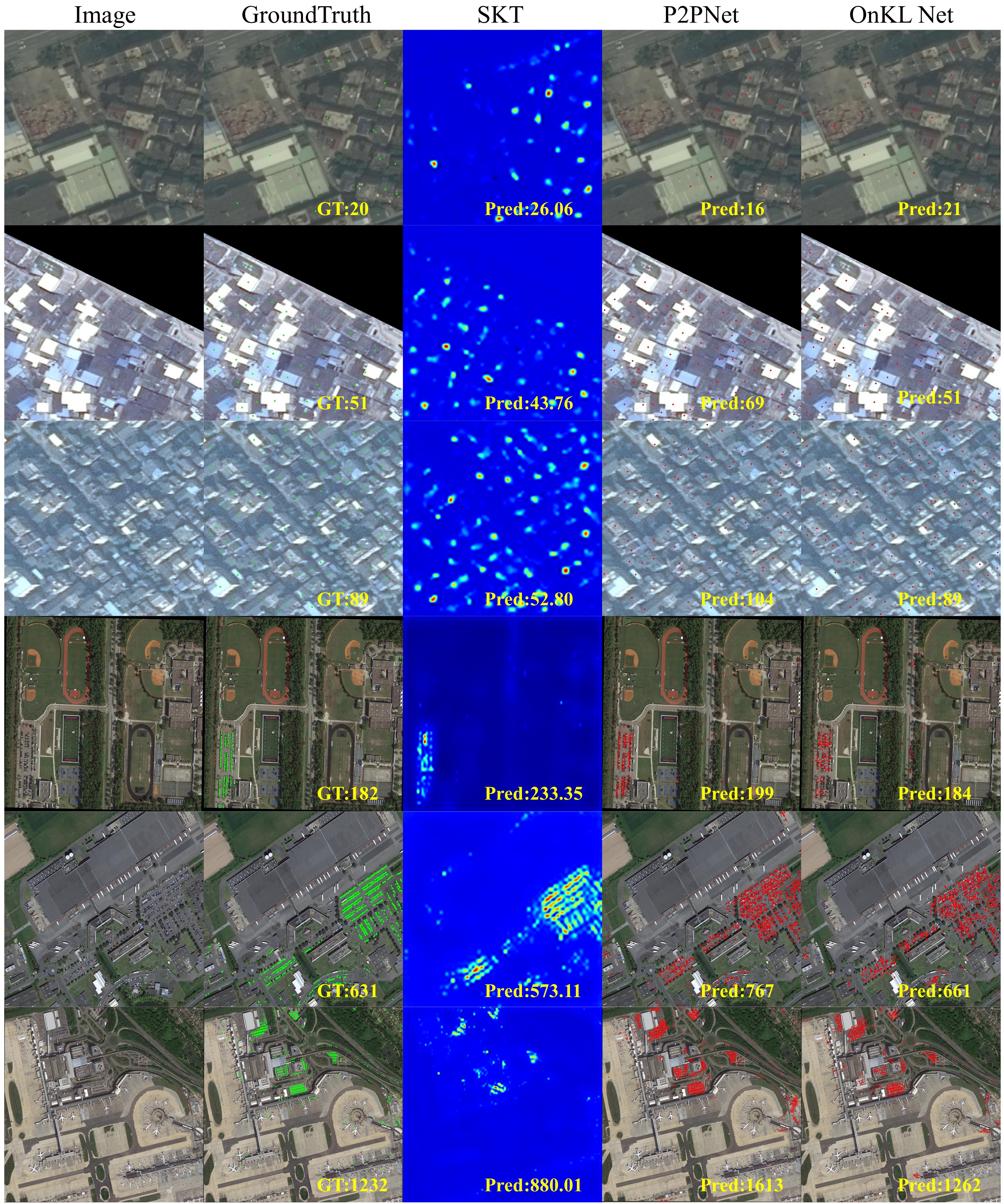}
	\caption{Comparison of predicted results on Building and Small-vehicle. {The first three rows show the images sampled from Building, while the remaining rows feature images of Small-vehicle.}}
	\label{fig:vis}
\end{figure*}

Furthermore, Fig.~\ref{fig:fig4} depicts a visualization of the inter-layer relationship matrix, as presented in Eq.~\ref{eq3}. The first row of visualizations reveals that parts of relationship matrices between shallow and intermediate layer features are dispersed, whereas the association matrices between these features and deep layer features are highly concentrated. {This difference may be because shallow features and intermediate features capture basic characteristics and slightly more complex structural information of inputs, respectively. In contrast to deep semantic features, these relatively coarse features, which may contain some noise, lead to greater variation in the internal relationships between features. High-level features, on the other hand, capture more abstract semantic information, resulting in higher similarity within their internal relationships. This causes the feature tensors representing intra-layer relationships to have higher brightness values along the diagonal, with lower values elsewhere. Here, we apply element-wise multiplication to model the inter-layer relationships. This operation allows the model to selectively enhance or suppress feature responses in specific regions. Therefore, we can observe that $b_1$($b_2$) and $c_1$($c_2$) in the figure are relatively similar. Furthermore, we also notice that the relationship matrix of the student branch is very close to that of the teacher branch, indicating that RiRD can effectively help the student network better understand the hierarchical feature differences.}



\begin{table}[t]
	\caption{Comparison of different feature distillation losses. {FD loss denotes the feature distillation loss.}}
	\label{tab:tab4}
	\centering
	\begin{tabular}{ccc}
		\hline
		Transfer Configuration       & MAE  &RMSE\\
		\hline
		W/O FD loss             &15.42  &24.29\\
		\hline
		+ SSIM loss                       &15.23  &24.08\\
		+ Cos loss                        &12.56  &17.81\\
		+ MSE loss                        &\textbf{12.27}  &\textbf{16.70}\\
		
		\hline
	\end{tabular}
\end{table}

\paragraph{Effects of Feature Distillation}

In this study, we assess the impact of different distillation losses on the performance of the student network by comparing three common losses: SSIM loss, Cos loss~\cite{liu2020efficient} and MSE loss. As shown in Table~\ref{tab:tab4}, all three feature distillation methods  can improve the student network's performance compared to not using any feature distillation loss. We notice that SSIM loss provides the least improvement. This may be because it places excessive emphasis on structural information while neglecting pixel relationships. It can hinder the student network's ability to capture subtle differences, which is important for detecting weak and small targets in remote sensing images. Both Cos loss and MSE loss perform similarly in terms of MAE, but MSE loss shows better generalization capability. Therefore, we select MSE loss as the feature distillation loss.

\paragraph{Effects of Different Numbers of Feature Groups} 

{In relation-in-relation distillation, the number of feature groups $M$ plays a crucial role in determining the effectiveness of the distillation. In this paper, we set $M$ to 3, based on the fact that the first ten layers of VGG-16 have a total of 4 blocks, with the shallowest block serving as a common block. Thus, we investigate the impact of the number of feature groups on network performance, as presented in Table~\ref{tab:tab4x}. From the results, it is evident that using only the deepest feature $ t_1 $ fails to generate meaningful inter-feature relationships, leading to poor performance. However, as shallower features are incorporated, the network performance improves significantly, with the best results achieved when $ M = 3 $.}

\begin{table}[t]
	\caption{Comparison of different numbers of feature groups. {$t_i$ ($s_i$) denotes the $i$th feature group in teacher (student) branch.}}
	\label{tab:tab4x}
	\centering
	\begin{tabular}{ccccc}
		\hline
		$t_1$ ($s_1$) & $t_2$ ($s_2$) & $t_3$ ($s_3$) & MAE   & RMSE  \\
		\hline
		$\checkmark$          &             &             & 16.34 & 23.92 \\
		$\checkmark$          & $\checkmark$           &             & 14.73 & 18.40 \\
		$\checkmark$           & $\checkmark$           & $\checkmark$           & \textbf{12.27} & \textbf{16.70} \\
		\hline
	\end{tabular}
\end{table}
\begin{table}[t]
	\caption{Comparison of different pretrained models. OnKL Net* denotes that we remove the residual connections in the student network.}
	\label{tab:tab4y}
	\centering
	\begin{tabular}{c|c|cc}
		\hline
		Method                   & Backbone & MAE   & RMSE  \\
		\hline
		\multirow{3}{*}{P2PNet}  & ResNet-34 & 19.91 & 64.97 \\
		& ResNet-50 & \textbf{11.28} & 18.50 \\
		& VGG-16      & 11.61 & \textbf{16.94} \\
		\hline
		\multirow{3}{*}{P2PNet*} & ResNet-34 & 16.50  & 37.95 \\
		& ResNet-50 & \textbf{12.63} & 19.08 \\
		& VGG-16      & 12.89 & \textbf{17.50} \\
		\hline
		\multirow{3}{*}{OnKL Net}    & ResNet-34 & 19.66 & 40.46 \\
		& ResNet-50 & 19.11 & 27.90 \\
		& VGG-16      & \textbf{12.27} & \textbf{16.70} \\
		\hline
		\multirow{2}{*}{OnKL Net*}   & ResNet-34 & 18.06 & 33.55 \\
		& ResNet-50 & \textbf{14.98} & \textbf{27.30}\\
		\hline
	\end{tabular}
\end{table}

\paragraph{Effects of Different Pretrained Models}

{We further examine the performance of different pre-trained backbone networks within our distillation framework, as summarized in Table~\ref{tab:tab4y}. The results show that in terms of pre-trained ResNets, the deeper version performs better on P2PNet and P2PNet*. By comparison, the VGG-16-based counting model shows greater potential for application than the ResNet-based one when considering both evaluation metrics. We speculate that this is due to the significant down-sampling in the early layers of ResNet, which limits the network's ability to capture fine details in remote sensing scenes—an essential factor for accurate object counting. When directly applying ResNet-34 and ResNet-50 to our method, we do not achieve satisfactory results. To address this, we build the student network by removing the residual structures and retaining only the $3\times3$ convolutional layers based on the channel-reduced ResNets. Clearly, it results in a notable  improvement. This suggests that, while residual structures aid in optimizing deep networks, they may not be suitable for the student model in our proposed framework. We speculate that with small datasets, residual structures could lead to overfitting, which in turn degrades performance.}

\begin{table*}[ht] 
	\caption{Comparison of our method with other methods on RSOC.  DM denotes the density map.}
	\label{tab:result}
	\centering
	\begin{tabular}{c|c|c|c|c|cc|cc|cc|cc}
		\hline
		\multirow{2}{*}{{Type}} & \multirow{2}{*}{{Method}} &  \multirow{2}{*}{{Time(min))}} & \multirow{2}{*}{{Param(M)}} & \multirow{2}{*}{{FLOPs(G)}}   & \multicolumn{2}{c|}{{Building}} & \multicolumn{2}{c|}{{Small-vehicle}} & \multicolumn{2}{c|}{{Large-vehicle}} & \multicolumn{2}{c}{{Ship}} \\
		\cline{6-13}
		& & & & & {MAE} & {RMSE} & {MAE} & {RMSE} & {MAE} & {RMSE} & {MAE} & {RMSE} \\
		\hline
		\multirow{9}{*}{\makecell{DM-based \\ methods}} & MCNN~\cite{zhang2016single} & 129 & 0.13 & 1.38  &12.13 &17.35  & 488.45 & 1310.44 & 21.69 & 39.41 & 93.76 & 133.18 \\
		& SKT~\cite{liu2020efficient} & 459 & 1.02 & 1.71 & \textbf{7.12} & 11.46  & 309.05 & 1073.72 &42.43 & 68.42 & 106.15 &146.74 \\
		\cline{2-13}
		
		& CSRNet~\cite{li2018csrnet} & 196 & 16.26 & 27.07 & 7.18 & 10.53 & 345.03 & 1039.16 & 35.88 & 48.32 & 71.92 & 109.63  \\
		& Bayesian\cite{ma2019bayesian} & 63 &21.50 & 26.99  & 28.95 & 32.93 & 185.79 & 703.31 & 26.03 & 51.54 & 64.15 & 86.01  \\
		& ASPD\cite{gao2020counting} & 265 &22.70 & 37.96  &7.59 &10.66 &--- & --- &42.87 & 63.49 &216.27 &352.23 \\
		& MAN\cite{lin2022boosting} & 103 &40.41 & 27.13  &29.01 &32.98 &505.10 &1326.12 &62.78 &79.65 &\textbf{47.20} &\textbf{70.24}  \\
		& PET\cite{liu2023point} & 75 &20.91 & 28.42  &10.78 &13.96 &419.74 &1212.18 &36.39 &54.52 & 258.83 &396.39  \\
		& GCFL\cite{shu2023generalized} & 8 &21.51 & 27.01  &7.79 &10.74 & 222.40 & 768.83 & 16.30 &23.31 	&55.32 &77.05  \\
		& PML\cite{yan2023progressive} & 43 &16.26 & 21.65  &8.3 &12.78 &217.17 &652.98&32.15 & 48.2 &139.58 &172.17 \\
		\hline
		\multirow{3}{*}{\makecell{Point-based \\ methods}} & P2PNet\cite{song2021rethinking} & 76  &19.20 & 23.29 &7.46 &\textbf{10.34} &174.00 &666.44 & \textbf{11.61} & 16.94 & 62.00 & 76.43 \\
		& P2PNet$^{*}$ & 72 &8.99 & 19.70  &8.25 & 11.12 & \textbf{159.33} & 638.36 & 12.89 & 17.50  &56.63 & 81.78  \\
		& OnKL Net & 85  & 0.76 & 4.00 &7.58 &10.62 &169.86 & \textbf{613.30} & 12.27 &\textbf{16.70} & 81.80 &105.57 \\
		\hline

	\end{tabular}
\end{table*}

\begin{table*}[]
	\caption{Comparison of our method with other methods on STAR. DM denotes the density map.}
	\label{tab:result1}
	\centering
	\begin{minipage}{\textwidth}\centering
		\scalebox{0.95}{\begin{tabular}{c|c|c|cc|cc|cc|cc|cc|cc}
		\hline
		 \multirow{2}{*}{{Type}} & \multirow{2}{*}{{Method}} & \multirow{2}{*}{{Param(M)}} & \multicolumn{2}{c|}{{Airplane}} & \multicolumn{2}{c|}{{Wind   mill}} & \multicolumn{2}{c|}{{Lattice   tower}} & \multicolumn{2}{c|}{{Bridge}} & \multicolumn{2}{c|}{{Runway}} & \multicolumn{2}{c}{{Average}} \\
		\cline{4-15}
		& &                           & MAE          &RMSE        &MAE            & RMSE           & MAE              &RMSE             &MAE          & RMSE        & MAE         & RMSE       &MAE          &RMSE         \\
		\hline
		\multirow{4}{*}{\makecell{DM-based \\ methods}} & Bayesian\cite{ma2019bayesian}   & 21.50                     & 45.94         & 66.08        & 18.94          & 25.22          & 25.93            & 36.00            & 2.28         & 5.20        & 1.39         & 1.76        & 18.90        & 26.85        \\
		& MAN\cite{lin2022boosting}   & 40.41                     & 46.82         & 66.68        & 19.73          & 25.88          & 26.54            & 36.43            & 3.02         & 5.96        & 2.06         & 2.31        & 19.63        & 27.45        \\
		& PML\cite{yan2023progressive}    & 16.26                     & 65.66         & 77.98        & 2.68           & 9.26           & 12.15            & 24.00            & 3.63         & 5.38        & 5.46         & 5.94        & 17.92        & 24.51        \\
		& GCFL\cite{shu2023generalized}   & 21.51                     & 14.18         & 23.37        & 2.42           & 3.42           & 17.96            & 28.41            & 2.10         & 5.39        & \textbf{0.82}         & \textbf{1.11}        & 7.50         & 12.34        \\
		\hline 
		\multirow{3}{*}{\makecell{Point-based \\ methods}} 
		& P2PNet\cite{song2021rethinking}  & 19.20                     & 12.20         & 21.92        & \textbf{0.45}           & \textbf{0.67}           & 12.21            & 21.05            & 1.60         & 2.46        & 1.18         & 1.51        & 5.53         & \textbf{9.52}         \\
		& P2PNet*                 & 8.99                      & \textbf{12.17}         & 21.61        & 0.64           & 1.00           & \textbf{11.56}            & \textbf{20.75}            & 1.65         & 2.83        & 1.35         & 1.75        & \textbf{5.47}         & 9.59         \\
		& OnKL Net                    & 0.76                      & 12.93         & \textbf{18.07}        & 2.00           & 2.56           & 15.36            & 23.53            & \textbf{1.53}         & \textbf{2.39}        & 1.18         & 1.67        & 6.60          & 9.64   \\
		\hline     
	\end{tabular}}
	\end{minipage}
\end{table*}

\subsection{Comparison with SOTA Methods}

Table~\ref{tab:result} shows the comparison results with some state-of-the-art (SOTA) methods. {As shown, our method achieves highly competitive results on RSOC with only 0.76M parameters and 4.0G FLOPs.} P2PNet serves as a strong baseline. Our method performs comparably or even better than P2PNet in the Building, Small-vehicle, and Large-vehicle datasets. However, in the Ship dataset, our results are less favorable, likely due to the training samples containing only 88 images, which makes the model prone to overfitting with such limited data. Similarly, {While our method yields results comparable to the baseline, P2PNet, it uses just 4\% of the parameters and 17\% of the FLOPs.} Moreover, we compare our method against density map-based methods such as MAN~\cite{lin2022boosting}, GCFL~\cite{shu2023generalized}, and PML~\cite{yan2023progressive}. {We notice that SKT outperforms our method in terms of MAE on Building. We speculate that this is due to two reasons: Firstly, SKT has a small number of parameters but still more than those of our method, giving it a stronger capability to fit the current dataset. Secondly, SKT utilizes a fixed Gaussian distribution to characterize each target, which can help it better distinguish between foreground and background areas. This could be important in the noisy remote sensing scenarios for regressing these buildings. In contrast, our method needs to generate a series of proposals and predict the confidence of these proposals, undoubtedly increasing the difficulty for the model to identify these regions of interest.} The results show that these methods achieve competitive outcomes on one or two subdatasets of RSOC but perform poorly on others, indicating limited adaptability. In contrast, our method consistently delivers highly competitive results across multiple datasets. Additionally, our method focuses on predicting point positions as the network's regression objective, which is more intuitive but also more challenging than density map-based methods. 

{While our method achieves satisfactory performance with reduced model inference overhead, it does require more training time compared to most other methods. This is primarily due to the different training objectives, i.e., the point-based objective vs. the density map-based objective. However, when compared to our baselines, P2PNet and P2PNet*, the increase in training time is minimal. Notably, in contrast to the two-stage distillation method SKT~\cite{liu2020efficient}, our method requires less than one-fifth of its training time. This demonstrates that our method incurs only a marginal increase in training cost while delivering an efficient inference model.}

Fig.~\ref{fig:vis} qualitatively presents the comparative results of our method with SKT and P2PNet under different densities. As shown, our method achieves excellent performance across various scenarios, particularly in medium- to low-density scenes. We also observe that in high-density scenes, there is a slight deviation between our predicted outcomes and the actual results. This deviation is due to the inherent challenges of robust counting and localization in scenes with extremely high target density. Furthermore, in contrast to SKT, which is based on density maps, our method is capable of predicting the location information of targets, thereby providing more robust support for scene understanding.

{In addition, we evaluate our method in the STAR dataset. Note that it is used for scene graph generation in large-scale satellite imagery, with object localization annotated via bounding boxes. Our method focuses on  object counting, while generating point-based locations for object localization. Note that STAR evaluates models through an online test dataset; however, due to inconsistencies with its label format, we are unable to assess our method’s performance using the online version. Therefore, we use the validation set to evaluate our method. The results, shown in Table~\ref{tab:result1}, demonstrate that our method achieves highly competitive performance with significantly fewer parameters. Specifically, it outperforms recent SOTA density map-based methods, such as PML and GCFL. Moreover, our method provides the location coordinates of objects, which facilitates a better analysis and understanding of remote sensing objects. When compared with P2PNet and P2PNet*, although our method yields slightly weaker results on average, it requires only about 4\% and 8\% of their parameters, respectively. In summary, these results further validate the effectiveness of our method.}

\subsection{Failure Cases}

\begin{figure}[ht] 
	\centering
	\includegraphics[scale = 0.3]{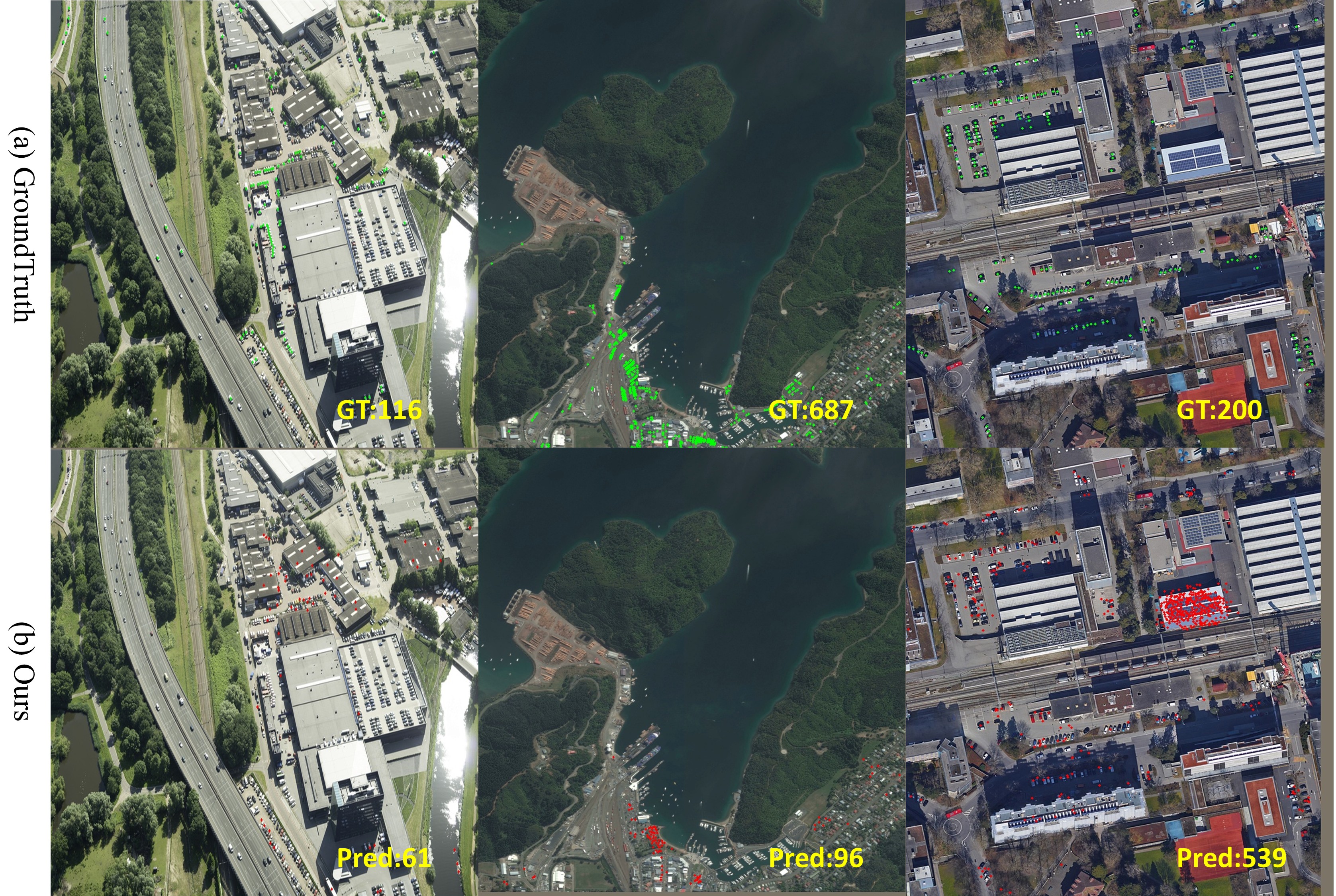}
	\caption{Examples of failed predictions of our network.}
	\label{fig:failure}
\end{figure}

{To better illustrate our method, we present three examples of failed predictions of our network in Fig.~\ref{fig:failure}. In the first column, we show the predicted results for large vehicles. The predicted value is 61, which is approximately half of the true count. This discrepancy can be attributed to the high-altitude angle from which the image was taken, reducing the apparent size of the target. Since object size is a key factor in recognition, it significantly complicates the accurate identification of vehicles. Additionally, certain background features, such as white rooftop decorations, closely resemble the targets, leading to potential misidentification by the model. The second column presents localization results for small vehicles. Due to the considerable distance between the camera and the targets, identifying these vehicles becomes more challenging. As shown, the predicted count is notably lower than the actual number. In the third scenario, we over-predict the number of small vehicles, highlighting the ongoing challenge of mitigating background interference in remote sensing tasks.}

\section{Conclusion}

In this study, we propose an efficient online distillation network for remote sensing object counting. Unlike the traditional two-stage distillation technique, we introduce an end-to-end online knowledge distillation framework. The framework comprises three components: a shared shallow module, a teacher branch, and a student branch, which integrates two traditionally distinct networks into a single trainable network. Then, we propose a new method for distilling feature relations, termed relation-in-relation distillation, which enables the student branch to better understand the evolution of inter-layer features. This is achieved by constructing an inter-layer relationship matrix that captures the relationship among inter-layer features.  Finally, our method achieves comparable results to several state-of-the-art models, with significantly fewer parameters, as demonstrated through extensive experiments on two challenging datasets. 

{While our method achieves competitive performance with fewer parameters, there’s still plenty of room for improvement. On one hand, since the student branch in our framework uses an empirical architecture, exploring automated architecture search strategies could help create a more lightweight and robust student network. On the other hand, exploring more efficient knowledge transfer strategies could further enhance the performance of the student branch.}

\bibliographystyle{IEEEbib}
\bibliography{bib_tip}


\end{document}